\documentclass[11pt]{article}
\usepackage{eacl2017}
\usepackage{times}
\usepackage{latexsym}
\usepackage{graphicx}
\usepackage{comment}
\usepackage{color}
\usepackage{paralist}
\usepackage{enumitem}
\usepackage{url}
\usepackage{xspace}
\usepackage{multirow}
\usepackage{cleveref}
\usepackage{booktabs}
\usepackage{amssymb}

\usepackage[utf8]{inputenc}

\newcommand{\wikid}{Wikidata\xspace}

\newcommand{\gend}{Genderize\xspace}
\newcommand{\avision}{AlchemyVision\xspace}
\newcommand{\av}{Alchemy\xspace}
\newcommand{\engfr}{\textit{en-fr}\xspace}
\newcommand{\freng}{\textit{fr-en}\xspace}
\newcommand{\engde}{\textit{en-de}\xspace}
\newcommand{\deeng}{\textit{de-en}\xspace}
\newcommand{\delang}{\textit{de}\xspace}
\newcommand{\englang}{\textit{en}\xspace}
\newcommand{\frlang}{\textit{fr}\xspace}
\newcommand{\roe}{ADD\xspace}

\newcommand{\human}{\textit{HT}\xspace}
\newcommand{\bl}{\textit{MT-B}\xspace}
\newcommand{\persthreem}{\textit{MT-P1}\xspace}
\newcommand{\perstwom}{\textit{MT-P2}\xspace}
\newcommand{\europarl}{EP\xspace}
\newcommand{\ted}{TED\xspace}

\usepackage{relsize}

\eaclfinalcopy 
\def\eaclpaperid{183} 

\newcommand{\note}[1]{\textit{\small\color{blue}{#1}}}
\setlist[itemize]{leftmargin=*}

\title{Personalized Machine Translation: Preserving Original Author Traits}

\author{
\fontsize{11}{10}\selectfont{
\centering
\begin{tabular}[t]{p{2cm}p{2cm}p{2cm}p{2cm}p{2cm}p{2cm}}
\multicolumn{2}{c}{Ella Rabinovich} & \multicolumn{2}{c}{Shachar Mirkin} & \multicolumn{2}{c}{Raj Nath Patel} \\
\multicolumn{2}{c}{\textnormal{Department of Computer Science}} & \multicolumn{2}{c}{\textnormal{IBM Research - Haifa}} & \multicolumn{2}{c}{\textnormal{C-DAC Mumbai}} \\
\multicolumn{2}{c}{\textnormal{University of Haifa, Israel}} & \multicolumn{2}{c}{\textnormal{Mount Carmel, Haifa}} & \multicolumn{2}{c}{\textnormal{Gulmohar Cross Road No. 9, Juhu}} \\
\multicolumn{2}{c}{\textnormal{\& IBM Research - Haifa}} & \multicolumn{2}{c}{\textnormal{31905, Israel}} & \multicolumn{2}{c}{\textnormal{Mumbai-400049, India}} \\
\multicolumn{2}{c}{\tt{ellarabi@gmail.com}} & \multicolumn{2}{c}{\tt{shacharm@il.ibm.com}} & \multicolumn{2}{c}{\tt{patelrajnath@gmail.com}} \\
& & & & & \\
\multicolumn{3}{c}{Lucia Specia} & \multicolumn{3}{c}{Shuly Wintner} \\
\multicolumn{3}{c}{\textnormal{Department of Computer Science}} & \multicolumn{3}{c}{\textnormal{Department of Computer Science}} \\
\multicolumn{3}{c}{\textnormal{University of Sheffield, United Kingdom}} & \multicolumn{3}{c}{\textnormal{University of Haifa, Israel}} \\
\multicolumn{3}{c}{\tt{l.specia@sheffield.ac.uk}} & \multicolumn{3}{c}{\tt{shuly@cs.haifa.ac.il}} \\
\end{tabular}
}
}

\date{}

\begin{document}
\maketitle

\begin{abstract}
The language that we produce reflects our personality, and various personal and demographic characteristics can be detected in natural language texts.
We focus on one particular personal trait of the author, \emph{gender}, and study how it is manifested in original texts and in translations. We show that author's gender has a powerful, clear signal in originals texts, but this signal is obfuscated in human and machine translation. We then propose simple domain-adaptation techniques that help retain the original gender traits in the translation, without harming the quality of the translation, thereby creating more personalized machine translation systems.

\end{abstract}

\section{Introduction}
\label{sec:intro}

Among many factors that mold the makeup of a text, gender and other authorial traits play a major role in our perception of the content we face.
Many studies have shown that these traits can be identified by means of automatic classification methods. Classical examples include gender identification \cite{KoppelAS02}, and authorship attribution and profiling \cite{Seroussi:2014:AAT}. Most research, however, addressed texts in a single language, typically English.

We investigate a related but different question: we are interested in understanding what happens to personality and demographic textual markers during the translation process. It is generally agreed that good translation goes beyond transformation of the original content, by preserving more subtle and implicit characteristics inferred by author's personality, as well as era, geography, and various cultural and sociological aspects. In this work we explore whether translations preserve the stylistic characteristic of the author and, furthermore, whether the prominent signals of the source are retained in the target language.

As a first step, we focus on \emph{gender} as a demographic trait (partially due to the absence of parallel data annotated for other traits). We evaluate the accuracy of automatic gender classification on original texts, on their manual translations and on their automatic translations generated through  statistical machine translation (SMT). We show that while gender has a strong signal in originals, this signal is obfuscated in human and machine translation. Surprisingly, determining gender over manual translation is even harder than over SMT; this may be an artifact of the translation process itself or the human translators involved in it.

Mirkin et al.~\shortcite{mirkin-EtAl:2015:EMNLP} were the first to show that authorial gender signals tend to vanish through both manual and automatic translation, using a small TED talks dataset. We use their data and extend it with a version of Europarl that we annotated with age and gender (\S \ref{sec:resource}). Furthermore, we conduct experiments with two language pairs, in both directions (\S\ref{sec:setting}). We also adopt a different classification methodology based on the finding that the translation process itself has a stronger signal than the author's gender (\S \ref{sec:evaluation-methodology}).

We then move on to assessing gender traits in SMT (\S\ref{sec:pers-models}).
Since SMT systems typically do not take personality or demographic information into account, we hypothesize that the author's style, affected by their personality, will fade. Furthermore, we propose simple domain-adaptation techniques that do consider gender information and can therefore better retain the original traits. We build ``gender-aware'' SMT systems, and show (\S\ref{sec:pers-proj-results}) that they retain gender markers while preserving general translation quality. Our findings therefore suggest that SMT can be made much more personalized, leading to translations that are more faithful to the style of the original texts. 

Finally, we analyze the prominent features that reflect gender in originals and translations (\S\ref{sec:pers-proj-analysis}).
Our experiments reveal that gender markers differ greatly by language, and the specific source language has a significant impact on the features and classification accuracy of the translated text. In particular, gender traits of the original language overshadow those of the target language in both manual and automatic translation products.

The \textbf{main contributions} of this paper are thus: {(i)} a new parallel corpus annotated with gender and age information, {(ii)} an in-depth assessment of the projection of gender traits in manual and automatic translation, and {(iii)} experiments showing that gender-personalized SMT systems better project gender traits while maintaining translation quality.


\section{Related work}
\label{sec:background}

While modeling of demographic traits has been proven beneficial in some NLP tasks such as sentiment analysis \cite{volkova-wilson-yarowsky:2013:EMNLP} or topic classification \cite{hovy:2015:ACL-IJCNLP}, very little attention has been paid to translation. We provide here a brief summary of research relevant to our work.

\paragraph{Machine translation (MT)}
Virtually no previous work in MT takes into account personal traits. State-of-the-art MT systems are built from examples of translations, where the general assumption is that the more data available to train models, the better, and a single model is usually produced. Exceptions to this assumption revolve around work on domain adaption, where systems are customized by using data that comes from a particular text domain \cite{hasler-haddow-koehn:2014,cuong2015latent}; and work on data cleaning, where spurious data is removed from the training set to ensure the quality of the final models \cite{cui-EtAl:2013:Short,AMTA-2014-Simard}. Personal traits, sometimes well marked in the translation examples, are therefore not explicitly addressed. Learning from different, sometimes conflicting writing styles can hinder model performance and lead to translations that are unfaithful to the source text. 

Focusing on reader preferences, Mirkin and Meunier~\shortcite{mirkin-meunier:2015:EMNLP} used a collaborative filtering approach from recommender systems, where a user's preferred translation is predicted based on the preferences of similar users. However, the user preferences in this case refer to the overall choice between MT systems of a specific reader, rather than a choice based on traits of the writer. Mirkin et al.~\shortcite{mirkin-EtAl:2015:EMNLP} motivated the need for personalization of MT models by showing that automatic translation does not preserve demographic and psychometric traits. They suggested treating the problem as a domain adaptation one, but did not provide experimental results of personalized MT models.

\paragraph{Gender classification}
A large body of research has been devoted to isolating distinguishing traits of male and female linguistic variations, both theoretically and empirically. Apart from content, male and female speech has been shown to exhibit stylistic and syntactic differences. Several studies demonstrated that literary texts and blog posts produced by male and female writers can be distinguished by means of automatic classification, using (content-independent) function words and n-grams of POS tags \cite{KoppelAS02,SchlerKAP06,Burger:2011}.

Although the tendencies of \textit{individual word} usage are a subject of controversy, distributions of \textit{word categories} across male and female English speech is nearly consensual: pronouns and verbs are more frequent in female texts, while nouns and numerals are more typical to male productions. Newman et al.~\shortcite{NewmanGHP:2008} carried out a comprehensive empirical study corroborating these findings with large and diverse datasets.

However, little effort has been dedicated to investigating the variation of individual markers of demographic traits across different languages. Johannsen et al.~\shortcite{johannsen:2015:CoNLL} conducted a large-scale study on linguistic variation over age and gender across multiple languages in a social media domain. They showed that gender differences captured by shallow syntactic features were preserved across languages, when examined by linguistic categories.
However, they did not study the distribution of individual gender markers across domains and languages. Our work demonstrates that while marker categories are potentially preserved, individual words typical to male and female language vary across languages and, more prominently, across different domains. 

\paragraph{Authorial traits in translationese}
A large body of previous research has established that translations constitute an autonomic \textit{language variety}: a special dialect of the target language, often referred to as \textit{translationese} \cite{Gellerstam:1986}. Recent corpus-based investigations of translationese demonstrated that originals and translations are distinguishable by means of supervised and unsupervised classification \cite{Baroni2006,vered:noam:shuly,TACL618}. The identification of machine-translated text has also been proven an easy task \cite{AraseZ13,AharoniKG14}.

Previous work has investigated how gender artifacts are carried over into human translation in the context of social and gender studies, as well as cultural transfer \cite{simon2003gender,von2010gender}. Shlesinger et al.~\shortcite{shlesinger2009markers} conducted a computational study exploring the implications of the translator's gender on the final product. They conclude that ``the computer could not be trained to accurately predict the gender of the translator''. Preservation of authorial style in literary translations was studied by Lynch~\shortcite{lynch2014supervised}, identifying Russian authors of translated English literature, by using (shallow) stylistic and syntactic features. Forsyth and Lam~\shortcite{forsyth2014found} investigated authorial discriminability in translations of French originals into English, inspecting two distinct human translations, as well as automatic translation of the same sources.

Our work, to the best of our knowledge, is the first to automatically identify speaker gender in manual, and more prominently, automatic translations over multiple domains and language-pairs, examining distribution of gender markers in source and target languages.


\section{Europarl with demographic info}
\label{sec:resource}

We created a resource\footnote{Available at \url{http://cl.haifa.ac.il/projects/pmt}} based on the parallel corpus of the European Parliament (Europarl) Proceedings \cite{Koehn05Europarl}. More specifically, we utilize the extension of its \engfr and \engde parallel versions \cite{rabinovich2015haifa}, where each sentence-pair is annotated with speaker name, the original language the sentence was uttered in, and the date of the corresponding session protocol. To extend speaker information with demographic properties, we used the Europarl website's MEP information pages\footnote{\url{http://www.europarl.europa.eu/meps/en/}} and applied a procedure of gender and age identification, as further detailed in \S \ref{sec:gender-prediction}.

The final resource comprises \engfr and \engde parallel bilingual corpora where metadata of members of the European Parliament (MEPs) is enriched with their gender and age at the time of the corresponding session. The data is restricted to sentence-pairs originally produced in English, French, or German. Table~\ref{tab:eu-gender-stats} provides statistics on the two datasets. We also release the full list of $3,586$ MEPs with their meta information. 

\begin{table}[hbt]
\centering
\begin{tabular}{p{2cm}|rr|rr}
& \multicolumn{1}{c}{\engfr} & \multicolumn{1}{c|}{\freng} & \multicolumn{1}{c}{\engde} & \multicolumn{1}{c}{\deeng} \\ \hline
male    & 100K  & 67K   & 101K  & 88K   \\
female  & 44K   & 40K   & 61K   & 43K   \\ \hline
total   & 144K  & 107K  & 162K  & 131K  \\
\end{tabular}
\caption{Europarl corpora (\europarl) statistics (\# of sentence-pairs); gender refers to an author of the source utterance.}
\label{tab:eu-gender-stats}
\end{table}

\subsection{Identification of MEP gender}
\label{sec:gender-prediction}
Gender annotation was conducted using three different resources: \wikid, \gend\ and \avision, which we briefly describe below.

\paragraph{\wikid}
\cite{wikidata} 
is a human-curated knowledge repository of structured data from Wikipedia and other Wikimedia projects. \wikid provides an API\footnote{\url{https://www.mediawiki.org/wiki/Wikibase/API}
} through which one can retrieve details about people in the repository, including place and date of birth, occupation, and gender. For MEPs found in the \wikid, we first verified that the person holds (or held) a position of {Member of the European Parliament} and if so, retrieved the gender. \wikid\ information is not complete: not all MEP names, positions or gender data is included. In total we obtained gender information for $2,618$ MEPs (73\% of the total $3,586$), of which $1,882$ (72\%) are male and $736$ female (28\%).

\paragraph{\gend\footnote{\url{https://genderize.io/}}}
 is an open resource containing over 2 million distinct names grouped by countries. It determines people's gender based on their first name and the country of origin. Provided with the first name and the country a MEP represents.\footnote{We assume that the country MEPs represent is highly correlated, if not strictly identical, to their country of origin.} \gend was able to predict the gender of $2,785$ MEPs,  the vast majority of them with a probability of $0.9$ or higher. We filtered out the $55$ lower-confidence entries, keeping $2,730$ MEPs (76\% of total), of which $2001$ (73\%) are male and $729$ (27\%) female.

\paragraph{\avision}
The European Parliament website maintains a page for every MEP, including personal photos. We classified MEP personal images using \avision,\footnote{\url{https://www.ibm.com/smarterplanet/us/en/ibmwatson/developercloud/alchemy-vision.html}} a publicly available image recognition service.
In total, we retrieved the gender of $2,236$ MEPs using \avision. Similarly to \gend, we filtered out all predictions with a confidence score below $0.9$, thus obtaining the gender of $2,138$ MEPs (60\% of total), of which $1,528$ are male and $610$ female ($71$\% and $29$\%, respectively).

\subsection{Resource evaluation and statistics}
\label{sec:stats-and-assessment}
Even though \wikid\ was created manually, to verify its correctness, we manually annotated the gender of $100$ randomly selected MEPs with available \wikid\ gender information; we found the metadata perfectly accurate. We therefore rely on \wikid\ as a gold-standard against which we can assess the accuracy of the two other resources. 
Table~\ref{tab:prediction-performance} presents the accuracy and coverage of each resource based on this methodology.

\begin{table}[hbt]
\centering
\resizebox{\columnwidth}{!}{
\begin{tabular} {l|r|r|r}
resource   & \wikid    & \gend     & \av \\ \hline
coverage   & 73.0      & 76.1      & 59.6 \\
accuracy   & 100.0     & 99.6      & 99.1\\
\end{tabular}
}
\caption{Gender prediction performance (\%).}
\label{tab:prediction-performance}
\end{table}

Given information obtained from the three resources, we assign each MEP with a single gender prediction in the following way: whenever it is found in \wikid\ ($2,618$ MEPs), the gender is determined by this resource. Otherwise, if both \gend\ and \avision\ produced agreed-upon gender information ($336$ out of $338$ cases), we set gender according to this prediction; the same applies to the case where only one of \gend\ or \avision\ provided a prediction ($346$ and $178$, respectively). We ended up with gender annotation for a total of $3,478$ out of $3,586$ members. The remaining $108$ MEPs ($92$ male, $16$ female) were annotated manually, a rather labor-intensive annotation in this case.

In total, the resource includes $947$ ($26$\%) female and $2,639$ ($74$\%) male MEPs. Based on the above accuracy estimations, and assuming that manual annotation is correct, the overall accuracy of gender information in this resource is $99.88$\%.

Utilizing the information on session dates and MEPs dates of birth available in the metadata, we also annotated each sentence-pair with the age of the MEP at the time the sentence was uttered. To summarize, we release the following resources:
\begin{inparaenum}[(i)]
\item meta information for $3,586$ MEPs, as described above,
\item bilingual parallel \engfr and \engde corpora, where each sentence-pair metadata is enriched with speaker MEPID, gender and age.
\end{inparaenum}


\section{Experimental setup}
\label{sec:setting}

We evaluate the extent to which gender traits are preserved in translation by evaluating the accuracy of gender classification of original and translated texts. The rationale is that the more prominent gender markers are in the text, the easier it is to classify the gender of its author.
\subsection{Translationese vs.\ gender traits}
\label{sec:evaluation-methodology}

\begin{figure}[hbt]
\center
\includegraphics[width=0.485\textwidth]{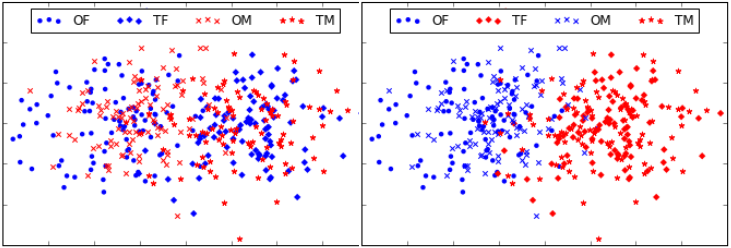}
\caption{English \europarl data distributions across two dimensions: gender (left) and trans. status (right).}
\vspace{-1em}
\label{fig:data-dist}
\end{figure}

Since we use the accuracy of gender identification as our evaluation metric, we isolate the dimension of gender in our data: the classification experiments are carried out separately on original, human translated text, as well as on each one of the MT products.
Human, and more prominently, machine translations constitute distinct and distinguishable language variation, characterized by unique feature distributions (\S \ref{sec:background}). We posit that in both human and machine translation products, the differences between original texts and translations overshadow the differences in gender.
We corroborate this assumption by analysing a sample data distribution by two dimensions: (i) translation status and (ii) gender. Figure~\ref{fig:data-dist} presents the results for the English Europarl corpus. Both charts display data distributions of the same four classes: original (O) and translated (T) English\footnote{This experiment refers to English translated from French; other language-pairs exhibited similar trends.} by male (M) and female (F) speakers (OM, OF, TM, TF). For the sake of visualization, the dimension of function words feature vectors was reduced to $2$, using principal component analysis \cite{jolliffe2002principal}. The left graph depicts color-separation by gender (male vs.\ female), while the right one by translation status (original vs.\ translated). Evidently, the linguistic variable of translationese stands out against the weaker signal of gender.

\subsection{Datasets}
\label{sec:datasets}

In addition to the Europarl corpus annotated for gender (\S \ref{sec:resource}), we experimented with a corpus of TED talks (transcripts and translations): a collection of texts from a completely different genre, where demographic traits may manifest differently. Testing the potential benefits of personalized SMT models on these two very diverse datasets allows us to examine the robustness of our approach.
We used the TED gender-annotated data from Mirkin et al.~\shortcite{mirkin-EtAl:2015:EMNLP}.\footnote{Downloaded from \url{http://cm.xrce.xerox.com/}.} This corpus contains annotation of the speaker's gender included in the English-French corpus of the IWSLT 2014 Evaluation Campaign's MT track \cite{wit3}. We annotated $68$ additional talks from the development and test sets of IWSLT 2014, 2015 and 2016. 
Using the full set, we split the TED parallel corpora by gender to obtain sub-corpora of 140K and 43K sentence pairs for male and female speakers, respectively.

\begin{table*}[!hbt]
\centering
\begin{tabular}{l|l|ccc|cc|cc}
\multicolumn{1}{c|}{} & \multicolumn{1}{c|}{} & \multicolumn{3}{c|}{training} & \multicolumn{2}{c|}{tuning} & \multicolumn{2}{c}{test} \\ \hline
\multicolumn{1}{c|}{dataset} & \multicolumn{1}{c|}{language-pair} & \multicolumn{1}{c}{M} & \multicolumn{1}{c}{F} & \multicolumn{1}{c|}{\roe} &
\multicolumn{1}{c}{M} & \multicolumn{1}{c|}{F} & \multicolumn{1}{c}{M} & \multicolumn{1}{c}{F} \\ \hline
\multicolumn{1}{l|}{\europarl} &\multicolumn{1}{l|}{\engfr \& \freng} & 144K & 65K & 1.71M & 2K & 2K & 15K & 15K \\
&\multicolumn{1}{l|}{\engde \& \deeng} & 170K & 86K & 1.50M & 2K & 2K & 15K & 15K \\ \hline
\multicolumn{1}{l|}{\ted} & \multicolumn{1}{l|}{\engfr} & 117K & 21K & 138K & 2K & 2K & 20K & 20K \\
\end{tabular}
\caption{MT datasets split for train, tuning and test, after cleaning.}
\vspace{-1em}
\label{tab:smt-datasets}
\end{table*}

The sizes of the datasets used for training, tuning and testing of SMT models are shown in Table~\ref{tab:smt-datasets}. Relatively large test sets are used for evaluation of the MT results for the sake of reliable per-outcome gender classification (\S \ref{sec:evaluation-methodology}).

Although the size of the training/tuning/test sets in either direction for any language-pair is the same, their content is different. We use data in both translation directions (i.e., \engfr and \freng, or \engde and \deeng) for both SMT experiments. Out of these data, 2K and 15K sentence-pairs (for each gender) are held out for tuning and test, respectively, where they comply with the translation direction. That is, for \engfr experiments, tuning and test sets are sampled from the \engfr direction only and vice-versa. The additional bilingual data (\roe) for training the models comes from the gender-unannotated portion of Europarl (all but the gender-annotated sub-corpus detailed in \S \ref{sec:resource}) for the \europarl experiments, and from combining \ted's male and female data for the experiments with TED.

\subsection{Classification setting}
\label{sec:cls}
All datasets were split by sentence, filtering out sentence alignments other than one-to-one. For POS tagging, we employed the Stanford implementation\footnote{\url{http://nlp.stanford.edu/software/tagger.shtml}} with its models for English, French and German. We divided all datasets into chunks of approximately 1,000 tokens, respecting sentence boundaries, and normalized the values of lexical features by the actual number of tokens in each chunk. For classification, we used Platt's sequential minimal optimization algorithm \cite{Keerthi2001} to train support vector machine classifiers with the default linear kernel \cite{weka}. In all experiments we used (the maximal) equal amount of data from each category (M and F), specifically, 370 chunks for each gender.

Aiming to abstract away from content and capture instead stylistic and syntactic characteristics, we used as our feature set the combination of function words (FW)\footnote{We used the lists of function words available at\\ \url{https://code.google.com/archive/p/stop-words}.} and (the top-1,000 most frequent) POS-trigrams. We employ 10-fold cross-validation for evaluation of classification accuracy.

\subsection{SMT setting}
We trained phrase-based SMT models with Moses~\cite{moses}, an open source SMT system. KenLM~\cite{kenlm} was used for language modeling. We trained 5-gram language models with Kneser-Ney smoothing~\cite{Chen:1996}. The models were tuned using Minimum Error Rate Tuning (MERT)~\cite{MERT}. Our preprocessing included cleaning (removal of empty, long and misaligned sentences), tokenization and punctuation normalization. The Stanford tokenizer~\cite{manning-EtAl:2014:P14-5} was used for tokenization and standard Moses scripts were used for other preprocessing tasks. We used BLEU~\cite{BLEU} to evaluate MT quality against one reference translation.


\section{Personalized SMT models}
\label{sec:pers-models}
In order to investigate and improve gender traits transfer in MT, we devise and experiment with gender-aware SMT models. We demonstrate that despite their simplicity, these models lead to better preservation of gender traits, while not harming the general quality of the translations.

We treat the task of personalizing SMT models as a domain adaptation task, where the {\em domain} is the gender. We applied two common techniques: (i) gender-specific model components (phrase table and language model (LM)) and (ii) gender-specific tuning sets. These personalized configurations are further compared to a baseline model where gender information is disregarded, as described below. In all cases, we use a single reordering table built from the entire training set.

\paragraph{Baseline}
The baseline (\bl) system was trained using the complete parallel corpus available for a language-pair. The training set contained both gender-specific and unannotated data, but no distinction was made between them. A single translation model and a single LM were built, and the model was tuned using a random sample of 2K sentence-pairs from the mixed data dedicated for tuning, preserving, therefore, the gender distribution of the underlying dataset.

\paragraph{Personalized models}
These models use three datasets: male, female, and additional in-domain bilingual data. 
Two configurations were devised: \persthreem, a model with three phrase tables and three LMs trained on the three datasets; and \perstwom, where for each gender a phrase table and a language model were built using only the gender-specific data, as well as a general phrase table and LM. In both configurations, each of the two genderized model variants was tuned using the gender-specific tuning set. In order to evaluate the translation quality of a personalized model, we separately translated the male and female source segments, merged the outputs and evaluated the merged result. 



\section{Results}
\label{sec:pers-proj-results}
Recall that we use the accuracy of gender classification as a measure of the strength of gender markers in texts. We assessed this accuracy below on originals and (human and machine) translations. First, however, we establish that the quality of SMT is not harmed with our personalized models.

\paragraph{MT evaluation}
We trained a baseline (\bl) and two personalized models (\persthreem and \perstwom) for each language pair as detailed in \S \ref{sec:pers-models}. The BLEU scores of \engfr and \freng personalized models were $38.42$, $38.34$ and $37.16$, $37.16$, with the baseline models scoring $38.65$ and $37.35$, respectively. Similarly, for experiments with \engde and \deeng and the TED data, the baseline scores ($21.95$, $26.37$ and $33.25$) were only marginally higher than those of the personalized models ($21.65$, $21.80$; $26.35$, $26.21$; and $33.19$, $33.16$), with differences ranging from $0.02$ to $0.3$. Neither \persthreem nor \perstwom  was consistently better than the other. We conclude, therefore, that all MT systems are comparable in terms of general quality.

\paragraph{Classification accuracy}
Tables~\ref{tab:eu-results-en-fr} and~\ref{tab:eu-results-en-de} present the results of gender classification accuracy in original (O), human- (HT) and machine-translated texts in the \europarl corpus. Female texts are distinguishable from their male counterparts with $77.3\%$ and  $77.1\%$ accuracy for English originals, in line with accuracies reported in the literature \cite{KoppelAS02}. Classification of original French and German texts reach $81.4\%$ (Table~ \ref{tab:eu-results-en-fr}) and $76.1\%$ (Table~\ref{tab:eu-results-en-de}), respectively.

\begin{table}[!hbt]
\centering
\resizebox{\linewidth}{!}{
\begin{tabular}{l|cc|cc|c}
\multicolumn{1}{c|}{} & \multicolumn{2}{c|}{precision} & \multicolumn{2}{c|}{recall} & acc. \\ \hline
dataset             & M & F & M & F & \\ \hline
\englang O          & 77.7 & 76.9 & 76.5 & 78.1 & 77.3 \\
\frlang O           & 80.9 & 81.9 & 82.2 & 80.5 & 81.4 \\ \hline
\freng \human       & 75.6 & 74.4 & 73.8 & 76.2 & 75.0 \\
\freng \bl          & 77.0 & 78.2 & 78.6 & 76.5 & 77.6 \\
\freng \persthreem  & 82.0 & 80.7 & 80.3 & 82.4 & \textbf{81.4} \\
\freng \perstwom    & 79.1 & 81.0 & 81.6 & 78.4 & 80.0 \\ \hline
\engfr \human       & 56.6 & 56.4 & 55.7 & 57.3 & 56.5 \\
\engfr \bl          & 60.2 & 60.1 & 60.0 & 60.3 & 60.1 \\
\engfr \persthreem  & 62.7 & 63.0 & 63.5 & 62.2 & 62.8 \\
\engfr \perstwom    & 65.2 & 65.3 & 65.4 & 65.1 & \textbf{65.3} \\
\end{tabular}
}
\caption{\europarl \engfr, \freng classification scores (\%).}
\label{tab:eu-results-en-fr}
\end{table}

\begin{table}[!hbt]
\centering
\resizebox{\linewidth}{!}{
\begin{tabular}{l|cc|cc|c}
\multicolumn{1}{c|}{} & \multicolumn{2}{c|}{precision} & \multicolumn{2}{c|}{recall} & acc. \\ \hline
dataset & M & F & M & F & \\ \hline
\englang O          & 77.5 & 76.7 & 76.5 & 77.7 & 77.1 \\
\delang O           & 76.4 & 75.7 & 75.4 & 76.8 & 76.1 \\ \hline
\deeng \human       & 68.6 & 67.9 & 67.3 & 69.2 & 68.2 \\
\deeng \bl          & 69.3 & 69.9 & 70.3 & 68.9 & 69.6 \\
\deeng \persthreem  & 77.4 & 75.9 & 75.1 & 78.1 & \textbf{76.6} \\
\deeng \perstwom    & 76.2 & 75.7 & 75.4 & 76.5 & 75.9 \\ \hline
\engde \human       & 59.8 & 59.7 & 59.5 & 60.0 & 59.7 \\
\engde \bl          & 63.8 & 64.0 & 64.3 & 63.5 & 63.9 \\
\engde \persthreem  & 69.6 & 69.4 & 69.2 & 69.7 & \textbf{69.5} \\
\engde \perstwom    & 66.7 & 67.7 & 68.6 & 65.7 & 67.2 \\
\end{tabular}
}
\caption{\europarl \engde, \deeng classification scores (\%).}
\label{tab:eu-results-en-de}
\end{table}

Evidently, gender traits are significantly obfuscated by both manual and non-personalized machine translation. The relatively low accuracy for human translation can be (partially) explained by the extensive editing procedure applied on Europarl proceedings prior to publishing \cite{cucchi:2012}, as well as the potential ``fingerprints'' of (male or female) human translators left on the final product.

Both \persthreem and \perstwom models yield translations that better preserve gender traits, compared to their manual and gender-agnostic automatic counterparts: accuracy improvements vary between $3.8$ for \freng translations to $7.0$ percent points for \deeng\footnote{All differences between \persthreem and \perstwom and baseline models are statistically significant.} (\persthreem vs \bl in both cases). Per-class precision and recall scores do not exhibit significant differences, despite the unbalanced amount of per-gender data used for training the MT models.

Gender classification results in the \ted dataset are presented in Table~\ref{tab:ted-results-en-fr}. The classification accuracy of English originals is $80.4\%$. While, similarly to Europarl, the gender signal is generally weakened in human translations\footnote{\ted talks are subtitled, rather than transcribed, undergoing some editing and rephrasing.} 
and baseline MT, overall accuracies are in most cases higher than in Europarl across all models. We attribute this difference to the more emotional and personal nature of TED speeches, compared with the formal language of the \europarl proceedings. Both personalized SMT models significantly outperform their baseline counterpart, as well as the manual translation, yielding $77.2\%$ and $77.7\%$ accuracy for \persthreem and \perstwom, respectively.

\begin{table}[!hbt]
\centering
\resizebox{\linewidth}{!}{
\begin{tabular}{l|cc|cc|c}
\multicolumn{1}{c|}{} & \multicolumn{2}{c|}{precision} & \multicolumn{2}{c|}{recall} & acc. \\ \hline
dataset             & M & F & M & F & \\ \hline
\englang O          & 81.2 & 79.7 & 79.2 & 81.6 & 80.4 \\ \hline
\engfr \human       & 74.0 & 73.5 & 73.2 & 74.3 & 73.8 \\
\engfr \bl          & 71.3 & 70.1 & 69.2 & 72.2 & 70.7 \\
\engfr \persthreem  & 77.5 & 76.8 & 76.5 & 77.8 & 77.2 \\
\engfr \perstwom    & 78.2 & 77.2 & 76.8 & 78.6 & \textbf{77.7} \\
\end{tabular}
}
\caption{\ted \engfr classification scores (\%).}
\label{tab:ted-results-en-fr}
\end{table}

\section{Analysis}
\label{sec:pers-proj-analysis}
\paragraph{Analysis of gender markers}
To analyze the extent to which personal traits are preserved in translations, we extract the set of most discriminative FWs in various texts by employing the InfoGain feature selection procedure \cite{gray1990entropy}. Gender markers vary across \textit{original} languages (with few exceptions); in \europarl, the most discriminating English features are \textit{also, very, perhaps, as, its, others, you}. The French list includes \textit{on, vous, dire, afin, doivent, doit, aussi, avait, voilà, je}, while the German list consists of \textit{wir, man, wirklich, sollten, von, für, dass, allen, ob}. The list of discriminative markers in the \ted\ English dataset contains mainly personal pronouns: \textit{she, her, I, you, my, our, me, and, who, it}.

\begin{figure*}[!hbt]
\center
\includegraphics[width=1\textwidth]{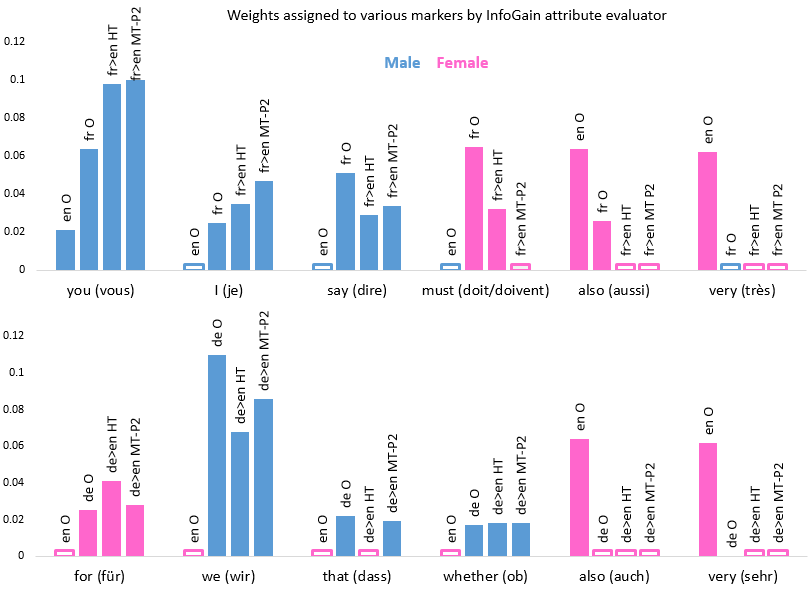}
\caption{Persistence of \englang and \frlang markers in \freng translations (top); \englang and \delang markers in \deeng translations (bottom). The transparent bars refer to (weak) F/M markers, assigned weight\textless $0.01$ by InfoGain.}
\label{fig:markers-fr-de-en}
\end{figure*}

Figure~\ref{fig:markers-fr-de-en} (top) presents weights assigned to various gender markers by the InfoGain attribute evaluator in originals and translations. Gender markers are carried over to (both manual and machine) translations to an extent that overshadows the original markers of the target language. In particular, the markers observed in translated English mirror their original French counterparts, in the same marker role: \textit{I} (M) in English translations reflecting the original French \textit{je} (M), \textit{say} (M) reflecting \textit{dire} (M), \textit{must} (F) translated from \textit{doit} (F) and \textit{doivent} (F); the latter contradicting the original English \textit{must} which characterizes M speech. The original English prominent gender markers (e.g., \textit{also, very}) almost completely lose their discriminative power in translations. A similar phenomenon is exhibited by English translations from German, as depicted in Figure~\ref{fig:markers-fr-de-en} (bottom): the German \textit{wir (we), für (for)} and \textit{ob (whether)} are preserved in (both manual and machine) English translations, in the same marker role.

We conclude that (i) gender traits in translation are weakened, compared to their originals. Furthermore, (ii) translations tend to embrace gender tendencies of the original language, thus resulting in a \textit{hybrid} outcome, where male and female traits are affected both by markers of the source and (to a much lesser extent) the target language.

\begin{table*}[!hbt]
\centering
\fontsize{10}{10}
\begin{tabular}{l|p{13cm}}
\frlang O & \textit{... on a corrigé la traduction du mot qui a été traduit en français par ``propriété'' qui n'est pas \textbf{vraiment} la même chose qu' ``appropriation''.
} \\ \hline
\freng \human & \textit{... it had been translated into French using the word for ``property'', which is not \textbf{really} the same thing as ``ownership''.} \\ \hline
\freng \bl & \textit{... it was corrected the translation of the word which has been translated into French as ``ownership'', which is not \textbf{really} the same as ``ownership''.} \\ \hline
\freng \persthreem & \textit{... it has corrected the translation of the word which has been translated into French as ``ownership'', which is not \textbf{exactly} the same as ``ownership''.} \\
\midrule[0.5mm]
\delang O & \textit{Entsprechend \textbf{halte ich es auch für notwendig}, daß die Kennzeichnung möglichst schnell und verpflichtend eingeführt wird, und zwar für Rinder und für Rindfleisch .
} \\ \hline
\deeng \human & \textit{Accordingly, \textbf{I consider it essential} that both the identification of cattle and the labelling of beef be introduced as quickly as possible on a compulsory basis.} \\ \hline
\deeng \bl & \textit{Similarly, \textbf{I believe that it is necessary}, as quickly as possible and that compulsory labelling will be introduced, and for bovine animals and for beef and veal.} \\ \hline
\deeng \persthreem & \textit{Accordingly, \textbf{I also think it is essential} that the labelling and become mandatory as quickly as possible, and for bovine animals and for beef.}
\end{tabular}
\caption{Translation of \frlang (M) and \delang (F) sentences into English manually, and by different MT models.}
\label{tab:trans-example-fr-de}
\end{table*}

\paragraph{Capturing the ``personalization'' effect}
Both manual- and all machine-translations of Europarl are tested on a strictly identical set of sentences; therefore, the performance gap introduced by personalized SMT models can be captured by a subset of sentences misclassified by the baseline model, but classified correctly when applying a more personalized approach. The inspection of differences in these translations can shed some light on the underlying nature of our personalized models. Table~\ref{tab:trans-example-fr-de} (top) shows manual, baseline, and personalized machine translations of examples of French and German sentences. The translation of the French word ``vraiment'' (in a \textit{male} utterance) varies in English as ``really'' or ``exactly'', where the former is more frequent in female English texts, and the latter is a male marker. The choice of a \textit{male} English marker over its \textit{female} equivalent by the gender-aware SMT model demonstrates the effect of personalization as proposed in this paper. The translations of the German \textit{female} sentence into English, as presented in Table~\ref{tab:trans-example-fr-de} (bottom), further highlight this phenomenon by choosing the English \textit{female} marker \textit{think} in its personalized translation over the more neutral \textit{consider} and \textit{believe} in the manual and baseline versions, respectively.

\section{Conclusions}
We presented preliminary results of employing personalized SMT models for better preservation of gender traits in automatic translation.
This work leaves much room for further research and practical activities. Authors' personal traits are utilized by recommendation systems, conversational agents and other personalized applications. While resources annotated for personality traits mainly exist for English (and recently, for a small set of additional languages), they are scarce or missing from most other languages. Employing MT models that are sensitive to authors' personal traits can facilitate user modeling in other languages as well as augment English data with translated content.

Our future plans include experimenting with more sophisticated MT models, and with additional demographic traits, domains and language-pairs.

\section*{Acknowledgments}
This research was partly supported by the H2020 QT21 project (645452, Lucia Specia).
We are grateful to Sergiu Nisioi for sharing the initial collection of properties of Members of the European Parliament.
We also thank our anonymous reviewers for their constructive feedback.

\bibliography{pmt16}
\bibliographystyle{eacl2017}

\end{document}